\documentclass[a4paper]{article}

\usepackage{INTERSPEECH2022}

\usepackage{hyperref}
\usepackage{color}
\definecolor{todoColor}{rgb}{1,0,1}

\def\mbf#1{\mathbf{#1}}

\def\bs#1{\boldsymbol{#1}}

\title{Self-supervised speech unit discovery \\from articulatory and acoustic features using VQ-VAE}

\name{Marc-Antoine Georges, Jean-Luc Schwartz, Thomas Hueber}
\address{
  Univ. Grenoble Alpes, CNRS, Grenoble INP, GIPSA-lab, 38000 Grenoble, France
 }
\email{marc-antoine.georges@grenoble-inp.fr}

\begin{document}

\maketitle
\begin{abstract}
  The human perception system is often assumed to recruit motor knowledge when processing auditory speech inputs. Using articulatory modeling and deep learning, this study examines how this articulatory information can be used for discovering speech units in a self-supervised setting. We used vector-quantized variational autoencoders (VQ-VAE) to learn discrete representations from articulatory and acoustic speech data. In line with the zero-resource paradigm, an ABX test was then used to investigate how the extracted representations encode phonetically relevant properties. Experiments were conducted on three different corpora in English and French. We found that articulatory information rather organises the latent representations in terms of place of articulation whereas the speech acoustics mainly structure the latent space in terms of manner of articulation. We show that an optimal fusion of the two modalities can lead to a joint representation of these phonetic dimensions more accurate than each modality considered individually. Since articulatory information is usually not available in a practical situation, we finally investigate the benefit it provides when inferred from the speech acoustics in a self-supervised manner.
\end{abstract}

\section{Introduction}

The discovery of phonological units 
from the speech input is a crucial stage in the development of infants and children. 
A long-standing question concerns the nature of phonetic cues enabling to establish and then access phonological units. Importantly, while much effort has been done to determine acoustic cues and possible acoustic/auditory invariants (e.g. \cite{stevens1980acoustic}), the role of articulatory knowledge has been proposed as critical in this process \cite{Liberman1985,browman1989articulatory}, particularly concerning the representation of consonantal features independently of their vowel context (the so-called ``search for invariance'' \cite{perkell2014invariance}).

The question of self-supervised unit discovery is also increasingly considered in the field of automatic speech processing. It is of potential great interest to design speech technology for low-resource languages  and it is one of the main tasks of the recent series of ``zero-resource challenges'' aiming at developing speech recognition and synthesis systems without textual resources \cite{dunbar2017zero, dunbar19_interspeech}. Still, most recent developments in this field only exploit the raw audio signal speech (e.g.  \cite{badino2014auto,lee2012nonparametric,synnaeve2014phonetics}) and the use of articulatory knowledge is rarely considered (or indirectly, by attempting to extract categorical articulatory or phonological features from sound, e.g.  \cite{baljekar15_interspeech}). This is mainly for practical rather than theoretical reasons. Articulatory data are generally not directly available in practical systems, and inferring them automatically from the audio speech signal of any arbitrary speaker (i.e. acoustic-to-articulatory inversion) is a difficult task.

\begin{figure}[t]
  \centering
  \includegraphics[width=\linewidth]{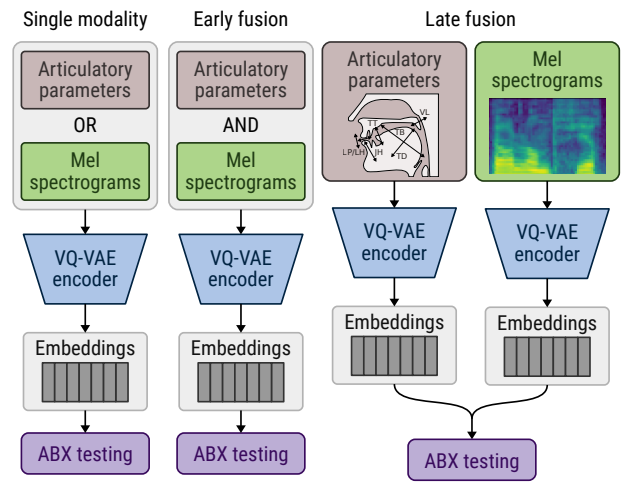}
   \caption{Proposed framework for discrete representation learning of speech from articulatory and acoustic speech data using VQ-VAE.}
  \label{fig:overview}
\end{figure}

In a series of previous studies, we have shown how introducing an explicit articulatory prior knowledge in the latent space of a variational auto-encoder (VAE) improves robustness to noise \cite{georges21_interspeech}. We have then proposed a self-supervised approach for learning a DNN-based acoustic-to-articulatory inversion model \cite{georges_et_al_icassp2022} trained end-to-end to ``repeat'' auditory speech inputs from any arbitrary speakers. In the present study, we specifically address the question of introducing articulatory knowledge in speech units discovery and focus on the question of invariance of consonants in varying vocalic context. In line with previous studies conducted in the scope of the zero-resource challenge \cite{niekerk20b_interspeech,Tjandra2019}, we use the VQ-VAE technique to learn quantized representational spaces likely to provide a basis for units representations. A set of VQ-VAEs were trained from ground truth articulatory data derived from available datasets for English and French, and compared with VQ-VAEs trained from their acoustic counterpart. For this comparison, we have used the ABX methodology \cite{schatz2013evaluating}. It assesses which linguistic invariants are encoded by the latent dimensions (in this framework, speech units can be defined from the emergence of sound categories well discriminated in the ABX paradigm). In line with the theoretical context mentioned previously, we focus on consonants, looking for representations that are robust to varying vocalic context. Then, we investigated how acoustic and articulatory modalities can be combined efficiently. Finally, as a first step toward a practical system, we replaced ground truth articulatory data by data inferred from the acoustic-to-articulatory inversion model proposed in our previous study  \cite{georges_et_al_icassp2022}.

\section{Material and methods}
\subsection{Datasets}

Two publicly available acoustic-articulatory datasets were used for this study. The first one is the Multichannel Articulatory Database (MOCHA) \cite{wrench2000multi}. Two speakers (fsew0, female and msak0, male) uttered 460 sentences extracted from the British TIMIT corpus, representing 20.6 minutes of speech for fsew0, and 17.4 minutes of speech for msak0.
For this corpus, each articulatory observation is a 14-dimensional vector gathering the 2D coordinates of 7 electromagnetic articulatory (EMA) coils describing the lips, tongue, jaw and velum positions in the midsagittal plane of each speaker's vocal tract, every 10~ms.

The second dataset is the PB2007 French dataset \cite{hueber_2015_taslp_cgmr} containing 1,109 items (15 minutes of speech) produced by a reference speaker (PB, male) \footnote{Dataset available at \url{https://doi.org/10.5281/zenodo.6390598}}. Here, each articulatory observation is a 12-dimensional vector (every 10~ms) gathering the 2D coordinates of six coils only compared to the MOCHA dataset, since the movements of the velum were not recorded.  An additional database of audio-only data was recorded by two other male speakers (TH and GB). Both were asked to pronounce the same corpus as the one recorded by the speaker PB.

\subsection{Articulatory and acoustic features}

In order to extract a set of dimensions which are interpretable in terms of articulatory control/function in speech production \cite{Maeda1994}, we applied the general methodology originally proposed by Maeda \cite{Maeda1990} and often referred to as the articulatory ``guided-PCA''. This linear dimensionality reduction technique aims at disentangling the movements of each speech articulator (e.g. by estimating the movements of the tongue and lips, independently of the jaw). To be concise, we do not recall here the details of this technique and we invite the reader to refer to \cite{georges21_interspeech} (section 2.2). 
After processing, the raw EMA data were reduced to a set of 7-dimensional feature vectors for the MOCHA dataset (1 for the jaw, 2 for the lips, 3 for the tongue, and 1 for the velum) and 6-dimensional feature vectors for the PB2007 dataset (same as MOCHA but with no velum).
For both datasets, a 40-dimensional mel-spectrogram was used to represent the acoustic content of the speech signal (extracted from the 16-kHz speech waveform recorded synchronously with the articulatory movements,  with a window size of 25~ms and a hop size of 10~ms using the Python \textit{librosa} toolkit).

\subsection{Vector Quantized Variational Autoencoder}

A vector quantized variational autoencoder (VQ-VAE) \cite{van2017neural} can be seen as a discrete version of the variational autoencoder (VAE) \cite{Kingma2014,rezende2014stochastic}. It has an encoder-decoder architecture, the decoder defining a posterior distribution $p_{\bs{\theta}}(\mbf{x} | \mbf{z})$ of the input observation $\mbf{x}$ given a latent variable $\mbf{z}$. The parameters of this distribution are provided by a DNN (with weights $\bs{\theta}$). Symmetrically, the encoder defines a posterior distribution $q_{\bs{\phi}}(\mbf{z} | \mbf{x})$ of the latent variable $\mbf{z}$ given an input observation $\mbf{x}$ (also parametrized by a DNN with weights $\bs{\phi}$). However, contrary to the VAE, this posterior distribution is categorical, such as:

\begin{equation}
q_{\bs{\phi}}(\mbf{z}=k \mid \mbf{x})= \begin{cases}1 & \text { for } \mathrm{k}=\operatorname{argmin}_{j}\left\|z_{e}(\mbf{x})-e_{j}\right\|_{2} \\ 0 & \text { otherwise }\end{cases}
\end{equation}
with $z_{e}(\mbf{x})$ the output of the encoder, and $\{e_{i}\}$ with ${i \in 1,2, \ldots, K}$ a set of $K$ $D$-dimensional latent embedding vectors. Therefore, in a VQ-VAE, the latent variable $\mbf{z}$ is discrete. It is defined as the index of the closest embedding vector w.r.t. the output of the encoder (nearest neighbour look-up). This embedding vector is then used as the input of the decoder. The set of embedding vectors (i.e. the codebook) is estimated from the data in a self-supervised manner, in addition to the parameters of both encoder and decoder neural networks.

\subsection{Implementation details}
\label{sec:implementation}
As illustrated in Figure \ref{fig:overview}, for each speaker and for each dataset, we trained one VQ-VAE  from articulatory data only, referred to as the ``articulatory VQ-VAE'' and a second one from the corresponding acoustic data (``acoustic VQ-VAE''). As a first way of combining both modalities, we investigated an ``early fusion'' strategy based on the concatenation of articulatory and acoustic feature vectors (resulting model is named ``articulatory-acoustic VQ-VAE'').
For each VQ-VAEs, the encoder was built with 3 fully connected layers of 256 neurons each, with the hyperbolic tangent used as the activation function (dropout and batch normalization layers were inserted after each fully connected layer with a dropout ratio of 0.25), and with a final linear layer of the size the dimension of the embedding vectors (i.e. $D$). A similar structure was used for the decoder, but with the final linear layer adapted to the size of the input data.

For each experiment, training data were z-scored. Model training was done using back-propagation with Adam optimizer, on mini-batches of 8 sequences of feature vectors. The implementation was done using the \textit{PyTorch} toolkit\footnote{The source code for the different experiments is available at: \url{https://gricad-gitlab.univ-grenoble-alpes.fr/georges1/articulatory-acoustic-vq-vae}} \cite{pytorch}.

For each experience, the datasets were randomly partitioned with 80\% of the data used for training and the remaining 20\% used for testing. 20\% of the training set was used as a validation set (for controlling the early stopping). For better robustness of the experimental results, each VQ-VAE model was trained and evaluated 5 times, with each time a different (random) partitioning of the datasets. We report here the average performance over these 5 evaluations.

\subsection{ABX evaluation}
\label{method:abx}

In line with the zero-resource challenge, the phonetic properties of the latent representations (i.e. the embedding vectors) learned by the different VQ-VAEs were assessed using a series of ABX tests. This method is based on the idea that the learned representation of two occurrences of the same phoneme (A and X) should be closer to each other than to any other phoneme (B). As stated previously, the study is focused on consonants in varied left and right vocalic contexts. For each dataset and speaker, and for each related test set, we extracted all vowel-consonant-vowel (VCV) sequences and their related (discrete) representations in latent spaces of all trained VQ-VAEs (articulatory, acoustic and articulatory-acoustic VQ-VAE). 
From these sequences, and for all consonants, we built a set of triplets A, B, X with A and X the representations associated with two occurrences of a same consonant but potentially in a different vocalic context, and B, the representations associated with an occurrence of a different consonant. We then compared the distance between A and X on one hand, and the distance between B and X on the other hand.

Here the distance is defined as the mean frame-wise cosine distance along a dynamic time warping (DTW) path, computed as follows: 1) for each VCVs (A, B and X), we extracted the sequence of corresponding embedding vectors using the different VQ-VAEs (articulatory, acoustic or articulatory-acoustic), 2) we kept only the frames corresponding to the central consonant (we rely here on the available segmentation at the phonetic level of PB2007 and MOCHA datasets), 3) we time-aligned the consonants of A and X on one hand, and B and X on the other hand using the DTW algorithm, 4) we calculated the average cosine distances  along the DTW path for both A vs. X ($d_{A,X}$) and B vs. X ($d_{B,X}$) (for this procedure, we used the toolkit provided in \cite{schatz2016-these}). A single ABX test is considered as a ``success'' when $d_{A,X}<d_{B,X}$.

To limit the computational cost, this ABX test was not done for all possible A, B, X triplets in each dataset, but only from a randomly selected subset of 5,000 (A, B, X) triplets ensuring that each (A, B) pair is evenly represented. For each VQ-VAE, and for each speaker, a global discriminability score was defined as the average success rate of all individual ABX tests.

In addition to the early-fusion strategy mentioned above based on the concatenation of articulatory and acoustic feature vectors and their modeling using a single VQ-VAE (as illustrated in Figure \ref{fig:overview}), we investigated another approach for combining the two modalities for the ABX tests.
When processing A and X stimuli, we computed $d^\textit{merge}_{A,X} = \omega . d^\textit{ac}_{A,X} + d^\textit{art}_{A,X}$ where $d^\textit{ac}_{A,X}$ is the distance between A and X obtained for the acoustic VQ-VAE (using the DTW-based procedure described above), and $d^\textit{art}_{A,X}$ is the distance between A and X but obtained with the articulatory VQ-VAE, and $\omega$ a weighing factor. We did the same for the (B, X) pair. A single ABX test was then considered as a ``success'' when $d^\textit{merge}_{A,X}<d^\textit{merge}_{B,X}$. This approach is referred here to as  ``late fusion''.

\subsection{Speech unit discovery from inferred articulatory data}

Since ground truth articulatory data are usually not available in a practical system, we investigated the use of an acoustic-to-articulatory inversion technique to recover them automatically from the raw speech signal. Acoustic-to-articulatory inversion is a well known regression problem which has been addressed with supervised    \cite{richmond_estimating_2002,toda_statistical_2008} or semi-supervised techniques \cite{hueber_2015_taslp_cgmr,girin_2017_taslp_jgmr}. In line with the zero-resource framework, we investigated for the present study the use of a self-supervised approach, and more specifically the one proposed in our previous study \cite{georges_et_al_icassp2022}. This approach is based on the coupling of a DNN-based acoustic-to-articulatory inversion model with a DNN-based articulatory-to-acoustic synthesis model, the latter being pre-trained in a supervised manner on articulatory-acoustic data of a reference speaker. The inversion model is trained to recover articulatory features in the vocal tract space of the reference speaker from audio-only inputs (potentially from any arbitrary speaker). The inferred articulatory features are then processed by the articulatory synthesizer which generates an audio output. The system is trained to minimize the discrepancy between audio inputs and corresponding synthetic audio outputs.  
For the present study, we pre-trained the articulatory synthesizer from the acoustic-articulatory data of the reference speaker PB in PB2007 dataset. We then trained the inversion model from the audio data of speakers TH and GB (from the same dataset), while freezing the parameters of the synthesis model. Both inversion and synthesis models were implemented as 4 fully connected layers with 256 neurons in each layer (with dropout and batch normalization technique used to improve generalization capability). We used the same train/validation/test partitioning as the one described in section \ref{sec:implementation}. More details about the architecture of the model can be found in \cite{georges_et_al_icassp2022}.

\section{Experiments and Results}

In order to calibrate the different VQ-VAE models used in the present study, we first conducted a series of preliminary simulations for assessing the impact of the number of embedding vectors (i.e. $K$) and their dimensionality (i.e. $D$). We did not observe significant improvements (in terms of overall ABX discriminability score) for $K\geq64$ and $D\geq32$. Thus, we report here only the results for models trained with these lower bounds (i.e. $K=64$ and $D=32$).

\subsection{Articulatory vs. acoustic VQ-VAEs}

\begin{figure}[t]
    \centering
    \includegraphics[width=\linewidth]{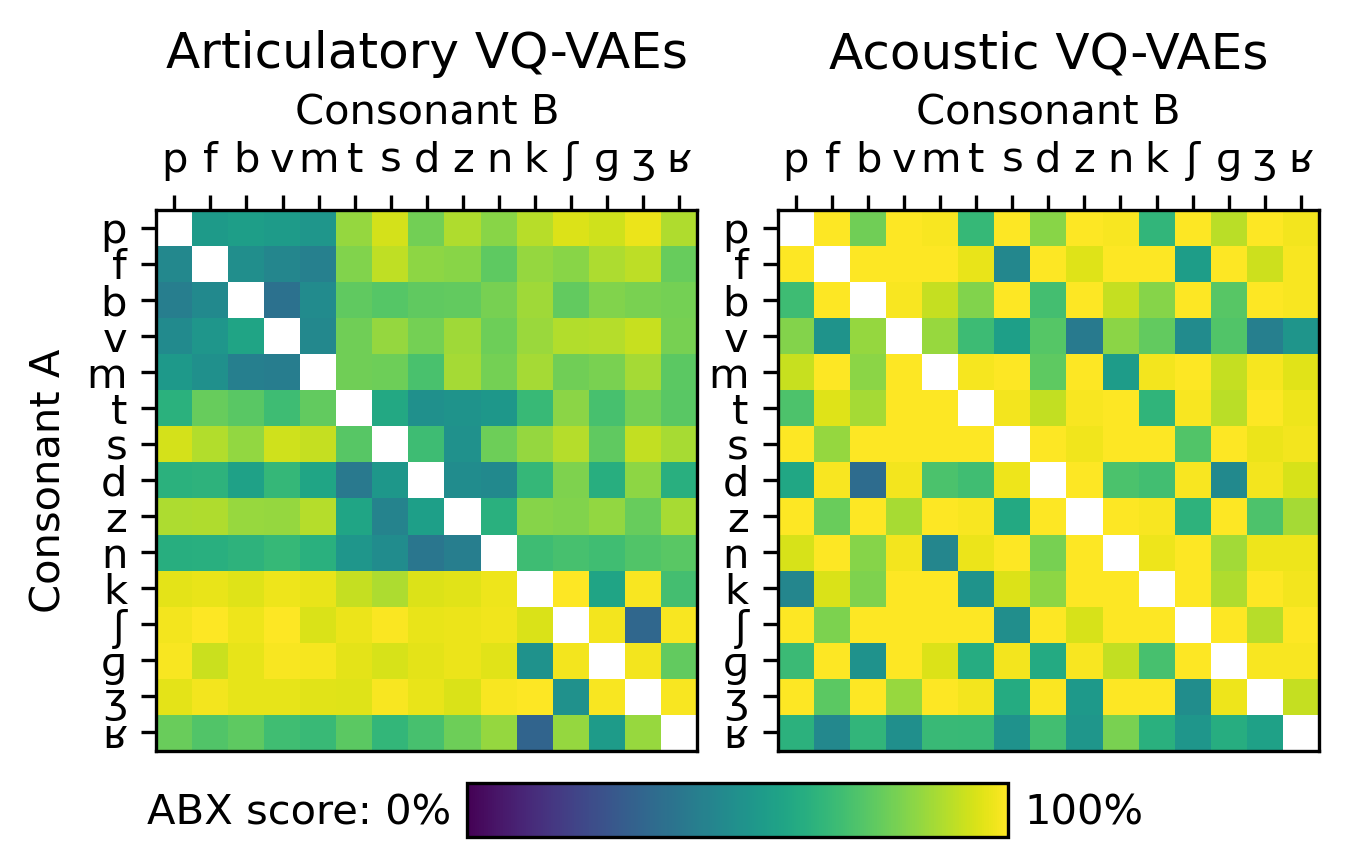}
    \caption{
        ABX discriminability scores for each pair of consonants computed for speaker PB (diagonal is empty because a consonant is never tested against itself).
    }
    \label{fig:matrix}
\end{figure}

For the 3 speakers PB, fsew0 and msak0, the ABX discriminability scores were respectively $86.4\%$, $78.4\%$, $78.8\%$ for the acoustic VQ-VAEs, and $77\%$, $76.7\%$, $78.3\%$ for the articulatory VQ-VAEs. Therefore, the acoustic and articulatory models behave differently only for the speaker PB. This may be explained by the lack of information about the velum for this speaker which makes it difficult for the learned representations to encode the nasality feature.

To better understand the error patterns for each modality, we report in Figure \ref{fig:matrix} the average ABX discriminability score for each pair of consonants, for speaker PB. We observe that for the articulatory VQ-VAE, the discriminability scores are lower for pairs of consonants with the same place of articulation (e.g. palatals [s] vs. [d] or labials [f] vs. [b]) whereas for the acoustic-VQ-VAE, scores are lower for pairs of consonants with the same manner of articulation (e.g. unvoiced fricatives [f] vs. [s]).

Then, we conducted a more fine-grained analysis in order to better investigate how the learned representations discriminate the consonants w.r.t. their place and manner of articulation. To that purpose, we defined two distinct ABX discriminability scores, respectively called ``manner of articulation ABX score'' and ``place of articulation ABX score''. To compute the first one (focusing on the manner of articulation), we grouped the consonants in three subgroups with similar place of articulation, that are labiodentals, palatals and dorsals. We applied the ABX testing methodology within each group (e.g. for the labiodental group, A is [abo], X is [iba] and B is [uvo]) and we calculated the success rate averaged over the three groups. To compute the second score (focusing on the place of articulation), we applied the same procedure but after grouping the consonant w.r.t. their manner of articulation. We considered the following five subgroups: voiced stop consonants, voiceless stop consonants, voiced fricatives/affricates, voiceless fricatives/affricates and sonorants i.e. liquids and nasals.

We reported these two scores for each speaker in Figure \ref{fig:fusion} (left column, blue and orange dots). The results confirm that articulatory information rather organises the latent representations in terms of place of articulation whereas the speech acoustics mainly structure the latent space in manner of articulation.
The different patterns between French and English speakers could be explained by the linguistic content of each dataset (mostly VCVs vs. sentences).

\subsection{Early vs. late fusion of the modalities}

\begin{figure}[t]
    \centering
    \includegraphics[width=.94\linewidth]{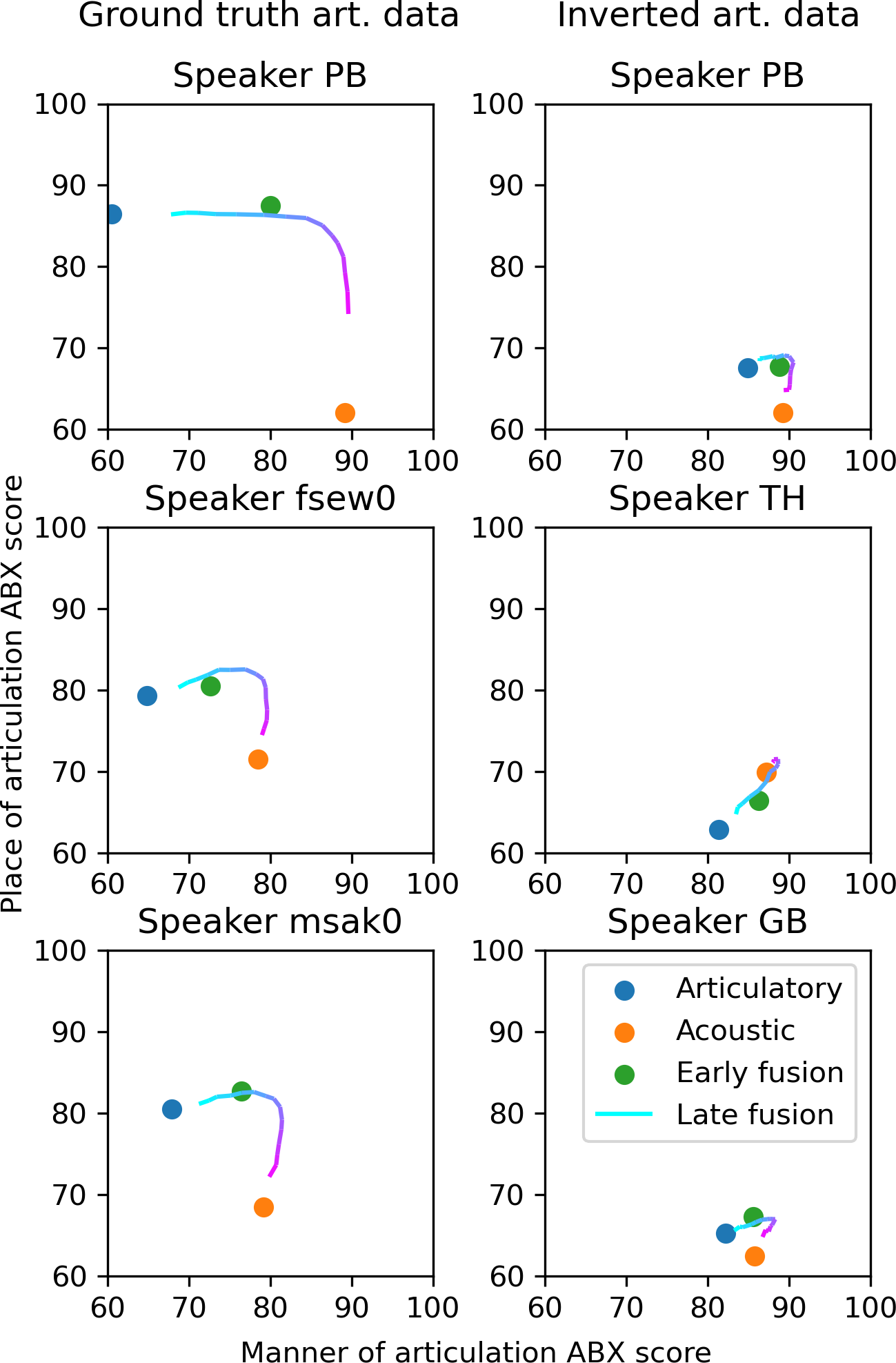}
    \caption{
        ABX scores w.r.t. place and manner of articulation for early and late fusion strategies, with ground-truth (left) and inverted (right) articulatory data.
        Late fusion performance varies with $\omega$ ranging from $10^{-1}$ (cyan) and $10^1$ (purple).
    }
    \label{fig:fusion}
\end{figure}

We reported also in Figure \ref{fig:fusion} the place vs. manner of articulation ABX discriminability scores, for both the early fusion (concatenation of acoustic and articulatory feature vectors, left column, green dots) and the late fusion strategies (combination of the ABX distances of the acoustic and articulatory VQ-VAEs, colored lines). The early fusion strategy gives slightly better results than the articulatory-only VQ-VAEs in term of place of articulation and a moderately weaker performance than the acoustic-only VQ-VAE in terms of manner of articulation.

As for the late fusion strategy, when modulating the contribution of both acoustic and articulatory VQ-VAE (i.e. $\omega$ varying between $10^{-1}$ and $10^1$), the performance follows a path with an optimum value that gives nearly as good performances as each modality taken independently for the PB2007 dataset and even better performances for the MOCHA dataset.
Therefore, these experiments tend to demonstrate that these fusion strategies can take advantage of both acoustic and articulatory modalities for speech unit discovery.

\subsection{Speech unit discovery from inferred articulatory data}

The experiments based on articulatory data inferred automatically from the speech audio signal were conducted on the PB2007 dataset only. The overall ABX discriminability scores for the speakers PB, GB and TH are also displayed on Figure \ref{fig:fusion} (right column).
First, the results obtained with speaker PB show that the representations are less structured in terms of place of articulation than the ones learned from ground truth articulatory data (see the ``place of articulation ABX scores'' of blue dots, comparing right and left top plots, $68\%$ vs. $86\%$). Nevertheless, for this speaker as well as for the speaker GB, we still see a lower 'manner of articulation ABX' score and higher 'place of articulation ABX score' for the articulatory VQ-VAEs than for the acoustic VQ-VAEs. Indeed to a lesser extent than when considering the ground truth articulatory data, we do also see an advantage to combining the two modalities.
However, for the TH speaker the acoustic score is already high (and much higher than for PB and GB) and as a consequence the inferred articulatory modality does not provide improvement.




\section{Conclusions}
We investigated the use of articulatory data for learning discrete representations of speech using VQ-VAEs.  Experiments conducted on both English and French languages, show a
potential importance of 
such data, at least for consonantal features. 
These results fit well with expectations about the role of articulatory features in the representation of consonant place of articulation \cite{Liberman1967}. 
The proposal of fusion strategies, either early or late, adds further light, suggesting that the complementarity of acoustic and articulatory representations could indeed be crucial for providing robust and complete phonetic representations. 



The passage to inferred articulatory input is less convincing. These representations, inferred in a completely self-supervised process driven by speech inputs from different speakers, are far from providing the complete information potentially available in the speaker’s articulatory knowledge. Since these inferred articulatory inputs actually provide rather efficient acoustic outputs \cite{georges_et_al_icassp2022}, this suggests that they are based on articulatory strategies more or less adequate in acoustic terms, but not sufficiently realistic in articulatory terms. The consequence is that VQ-VAEs based on these representations are actually close to acoustic VQ-VAEs. Future work will aim at extending the articulatory-to-acoustic self-supervised learning process in \cite{georges_et_al_icassp2022} to better exploit the present developments about the discovery of speech units, relating the imitation and unit discovery processes in a computational  model of sensory-motor speech learning.  

\section{Acknowledgements}

This work has been partially supported by MIAI @ Grenoble Alpes (ANR-19-P3IA-0003).
The authors would like to thank Julien Diard and Laurent Girin for fruitful discussions.

\newpage
\bibliographystyle{IEEEtran}

\bibliography{mybib}

\end{document}